\documentclass{article}

\usepackage{arxiv}

\usepackage[utf8]{inputenc} % allow utf-8 input
\usepackage[T1]{fontenc}    % use 8-bit T1 fonts
\usepackage{hyperref}       % hyperlinks
\usepackage{url}            % simple URL typesetting
\usepackage{booktabs}       % professional-quality tables
\usepackage{amsfonts}       % blackboard math symbols
\usepackage{nicefrac}       % compact symbols for 1/2, etc.
\usepackage{microtype}      % microtypography
\usepackage{lipsum}

\usepackage{graphicx}
\graphicspath{ {./images/} }

\title{Citation Data of Czech Apex Courts}

\author{
  Jakub~Harašta \\
  Institute of Law and Technology\\
  Masaryk University\\
  Brno, CZ \\
  \texttt{jakub.harasta@law.muni.cz} \\
   \And
 Tereza~Novotná \\
  Institute of Law and Technology\\
  Masaryk University\\
  Brno, CZ \\
  \texttt{tereza.novotna@mail.muni.cz} \\
   \AND
  Jarom\'ir~\v{S}avelka \\
  Intelligent Systems Program\\
  University of Pittsburgh\\
  Pittsburgh, USA \\
  \texttt{jas438@pitt.edu} \\
}

\begin{document}
\maketitle

\begin{abstract}
In this paper, we introduce the citation data of the Czech apex courts (Supreme Court, Supreme Administrative Court and Constitutional Court). This dataset was automatically extracted from the corpus of texts of Czech court decisions - CzCDC 1.0. We obtained the citation data by building the natural language processing pipeline for extraction of the court decision identifiers. The pipeline included the (i) document segmentation model and the (ii) reference recognition model. Furthermore, the dataset was manually processed to achieve high-quality citation data as a base for subsequent qualitative and quantitative analyses. The dataset will be made available to the general public.

%In this paper, we introduce the citation data of the Czech apex courts (Supreme Court, Supreme Administrative Court and Constitutional Court). We obtained the data by building the natural language processing pipeline for extracting court decision identifiers. The pipeline included the (i) document segmentation model and the (ii) reference recognition model. This layered approach allowed us to outperform the previously presented reference recognition model, and achieve the F1 measure of .815.

%Following the extraction of raw data, we further processed the data manually to achieve high-quality dataset as a base for subsequent qualitative and quantitative analyses. Overall, we have extracted 903 019 court decision identifiers from 237 723 decisions contained in the CzCDC 1.0 dataset. We have parsed these identifiers into 668 365 unique references (459 198 references within the CzCDC 1.0 dataset). The dataset will be made available to general public.

\end{abstract}

% keywords can be removed
\keywords{Reference recognition \and
Reference extraction \and Document segmentation \and NLP pipeline \and
Citation data \and Supreme Court \and Supreme Administrative Court \and Constitutional Court \and
Czech Republic}

\section{Introduction}
Analysis of the way court decisions refer to each other provides us with important insights into the decision-making process at courts. This is true both for the common law courts and for their counterparts in the countries belonging to the continental legal system. Citation data can be used for both qualitative and quantitative studies, casting light in the behavior of specific judges through document analysis or allowing complex studies into changing the nature of courts in transforming countries.

That being said, it is still difficult to create sufficiently large citation datasets to allow a complex research. In the case of the Czech Republic, it was difficult to obtain a relevant dataset of the court decisions of the apex courts (Supreme Court, Supreme Administrative Court and Constitutional Court). Due to its size, it is nearly impossible to extract the references manually. One has to reach out for an automation of such task. However, study of court decisions displayed many different ways that courts use to cite even decisions of their own, not to mention the decisions of other courts.The great diversity in citations led us to the use of means of the natural language processing for the recognition and the extraction of the citation data from court decisions of the Czech apex courts.

In this paper, we describe the tool ultimately used for the extraction of the references from the court decisions, together with a subsequent way of manual processing of the raw data to achieve a higher-quality dataset. Section \ref{sec:related} maps the related work in the area of legal citation analysis (Section\ref{sec:legcitana}), reference recognition (Section \ref{sec:refrec}), text segmentation (Section \ref{sec:docseg}), and data availability (Section \ref{sec:datava}). Section \ref{sec:methodology}
describes the method we used for the citation extraction, listing the individual models and the way we have combined these models into the NLP pipeline. Section \ref{sec:results} presents results in the terms of evaluation of the performance of our pipeline, the statistics of the raw data, further manual processing and statistics of the final citation dataset. Section \ref{sec:discussion} discusses limitations of our work and outlines the possible future development. Section \ref{sec:conclusion} concludes this paper.

\section{Related work}
\label{sec:related}

\subsection{Legal Citation Analysis}
\label{sec:legcitana}
The legal citation analysis is an emerging phenomenon in the field of the legal theory and the legal empirical research.The legal citation analysis employs tools provided by the field of network analysis.

In spite of the long-term use of the citations in the legal domain (eg. the use of Shepard's Citations since 1873), interest in the network citation analysis increased significantly when Fowler et al. published the two pivotal works on the case law citations by the Supreme Court of the United States \cite{fowler2007,fowler2008}. Authors used the citation data and network analysis to test the hypotheses about the function of \textit{stare decisis} the doctrine and other issues of legal precedents. In the continental legal system, this work was followed by Winkels and de Ruyter \cite{winkels}. Authors adopted similar approach to Fowler to the court decisions of the Dutch Supreme Court. Similar methods were later used by Derlén and Lindholm \cite{derlen-german, derlen-economic} and Panagis and Šadl \cite{panagis} for the citation data of the Court of Justice of the European Union, and by Olsen and Küçüksu for the citation data of the European Court of Human Rights \cite{olsen}.

Additionally, a minor part in research in the legal network analysis resulted in the past in practical tools designed to help lawyers conduct the case law research. Kuppevelt and van Dijck built prototypes employing these techniques in the Netherlands \cite{kuppevelt}. Görög a Weisz introduced the new legal information retrieval system, Justeus, based on a large database of the legal sources and partly on the network analysis methods. \cite{gorog}

\subsection{Reference Recognition}
\label{sec:refrec}
The area of reference recognition already contains a large amount of work. It is concerned with recognizing text spans in documents that are referring to other documents. As such, it is a classical topic within the AI \& Law literature.

The extraction of references from the Italian legislation based on regular expressions was reported by Palmirani et al. \cite{palmirani}. The main goal was to bring references under a set of common standards to ensure the interoperability between different legal information systems. 

De Maat et al. \cite{maat} focused on an automated detection of references to legal acts in Dutch language. Their approach consisted of a grammar covering increasingly complex citation patterns.

Opijnen \cite{opijnen} aimed for a reference recognition and a reference standardization using regular expressions accounting for multiple the variant of the same reference and multiple vendor-specific identifiers.

The language specific work by Kríž et al. \cite{kriz} focused on the detecting and classification references to other court decisions and legal acts. Authors used a statistical recognition (HMM and Perceptron algorithms) and reported F1-measure over 90\% averaged over all entities. It is the state-of-art in the automatic recognition of references in the Czech court decisions. Unfortunately, it allows only for the detection of docket numbers and it is unable to recognize court-specific or vendor-specific identifiers in the court decisions.

Other language specific-work includes our previous reference recognition model presented in \cite{harasta}. Prediction model is based on conditional random fields and it allows recognition of different constituents which then establish both explicit and implicit case-law and doctrinal references. Parts of this model were used in the pipeline described further within this paper in Section \ref{sec:methodology}.

\subsection{Data Availability}
\label{sec:datava}

Large scale quantitative and qualitative studies are often hindered by the unavailability of court data. Access to court decisions is often hindered by different obstacles. In some countries, court decisions are not available at all, while in some other they are accessible only through legal information systems, often proprietary. This effectively restricts the access to court decisions in terms of the bulk data. This issue was already approached by many researchers either through making available selected data for computational linguistics studies or by making available datasets of digitized data for various purposes. Non-exhaustive list of publicly available corpora includes British Law Report Corpus \cite{perez}, The Corpus of US Supreme Court Opinions \cite{davies},the HOLJ corpus \cite{grover}, the Corpus of Historical English Law Reports, Corpus de Sentencias Penales \cite{rodriguez}, Juristisches Referenzkorpus \cite{hamann} and many others.

Language specific work in this area is presented by the publicly available Czech Court Decisions Corpus (CzCDC 1.0) \cite{novotna}. This corpus contains majority of court decisions of the Czech Supreme Court, the Supreme Administrative Court and the Constitutional Court, hence allowing a large-scale extraction of references to yield representative results. The CzCDC 1.0 was used as a dataset for extraction of the references as is described further within this paper in Section \ref{sec:methodology}. Unfortunately, despite containing 237 723 court decisions issued between 1st January 1993 and 30th September 2018, it is not complete. This fact is reflected in the analysis of the results.

\subsection{Document Segmentation}
\label{sec:docseg}

A large volume of legal information is available in unstructured form, which makes processing these data a challenging task – both for human lawyers and for computers. Schweighofer \cite{schweighofer} called for generic tools allowing a document segmentation to ease the processing of unstructured data by giving them some structure.

Topic-based segmentation often focuses on the identifying specific sentences that present borderlines of different textual segments. 

The automatic segmentation is not an individual goal – it always serves as a prerequisite for further tasks requiring structured data. Segmentation is required for the text summarization \cite{barzilay, hearst}, keyword extraction \cite{ercan}, textual information retrieval \cite{prince}, and other applications requiring input in the form of structured data.

Major part of research is focused on semantic similarity methods.The computing similarity between the parts of text presumes that a decrease of similarity means a topical border of two text segments. This approach was introduced by Hearst \cite{hearst} and was used by Choi \cite{choi} and Heinonen \cite{heinonen} as well. 

Another approach takes word frequencies and presumes a border according to different key words extracted. Reynar \cite{reynar} authored graphical method based on statistics called dotplotting. Similar techniques were used by Ye \cite{ye} or Saravanan \cite{saravanan}. Bommarito et al. \cite{bommarito} introduced a Python library combining different features including pre-trained models to the use for automatic legal text segmentation. Li \cite{li} included neural network into his method to segment Chinese legal texts.  

\v{S}avelka and Ashley \cite{savelka} similarly introduced the machine learning based approach for the segmentation of US court decisions texts into seven different parts. Authors reached high success rates in recognizing especially the Introduction and Analysis parts of the decisions. 

Language specific work includes the model presented by Harašta et al. \cite{harasta-segm}. This work focuses on segmentation of the Czech court decisions into pre-defined topical segments. Parts of this segmentation model were used in the pipeline described further within this paper in Section \ref{sec:methodology}. 

\section{Methodology}
\label{sec:methodology}
%Jakub: máme dataset rozhodnutí; reference recognition model (JURIX 2017); segmentační model (Jusletter); pipeline k řešení některých nedostatků identifikovaných v error analýze výkonu modelu z JURIX 2017

In this paper, we present and describe the citation dataset of the Czech top-tier courts. To obtain this dataset, we have 
processed the court decisions contained in CzCDC 1.0 dataset by the NLP pipeline consisting of the segmentation model introduced in \cite{harasta-segm}, and parts of the reference recognition model presented in \cite{harasta}. The process is described in this section.

\subsection{Dataset and models}

\subsubsection{CzCDC 1.0 dataset}
Novotná and Harašta \cite{novotna} prepared a dataset of the court decisions of the Czech Supreme Court, the Supreme Administrative Court and the Constitutional Court. The dataset contains 237,723 decisions published between 1$^{st}$ January 1993 and the 30$^{th}$ September 2018. These decisions are organised into three sub-corpora. The sub-corpus of the Supreme Court contains 111,977 decisions, the sub-corpus of the Supreme Administrative Court contains 52,660 decisions and the sub-corpus of the Constitutional Court contains 73,086 decisions. Authors in \cite{novotna} assessed that the CzCDC currently contains approximately 91\% of all decisions of the Supreme Court, 99,5\% of all decisions of the Constitutional Court, and 99,9\% of all decisions of the Supreme Administrative Court. As such, it presents the best currently available dataset of the Czech top-tier court decisions.

\subsubsection{Reference recognition model}
Harašta and Šavelka \cite{harasta} introduced a reference recognition model trained specifically for the Czech top-tier courts. Moreover, authors made their training data available in the \cite{harasta-data}. Given the lack of a single citation standard, references in this work consist of smaller units, because these were identified as more uniform and therefore better suited for the automatic detection. The model was trained using conditional random fields, which is a random field model that is globally conditioned on an observation sequence \textit{O}. The states of the model correspond to event labels \textit{E}. Authors used a first-order conditional random fields. Model was trained for each type of the smaller unit independently.

\subsubsection{Text segmentation model}
Harašta et al. \cite{harasta-segm}, authors introduced the model for the automatic segmentation of the Czech court decisions into pre-defined multi-paragraph parts. These segments include the Header (introduction of given case), History (procedural history prior the apex court proceeding), Submission/Rejoinder (petition of plaintiff and response of defendant), Argumentation (argumentation of the court hearing the case), Footer (legally required information, such as information about further proceedings), Dissent and Footnotes. The model for automatic segmentation of the text was trained using conditional random fields. The model was trained for each type independently.

\subsection{Pipeline}
\label{pipeline}
\begin{figure}
    \centering
    \includegraphics[width=16cm, height=3cm]{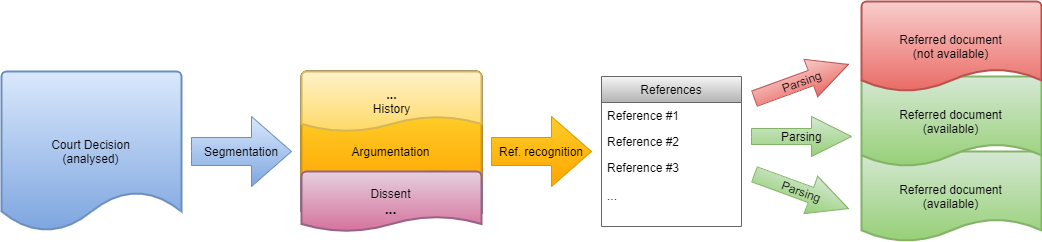}
    \caption{NLP pipeline including the text segmentation, reference recognition and parsing of references to the specific document}
    \label{fig:pipeline}
\end{figure}

In order to obtain the citation data of the Czech apex courts, it was necessary to recognize and extract the references from the CzCDC 1.0. Given that training data for both the reference recognition model \cite{harasta, harasta-data} and the text segmentation model \cite{harasta-segm} are publicly available, we were able to conduct extensive error analysis and put together a pipeline to arguably achieve the maximum efficiency in the task. The pipeline described in this part is graphically represented in Figure \ref{fig:pipeline}.

As the first step, every document in the CzCDC 1.0 was segmented using the text segmentation model. This allowed us to treat different parts of processed court documents differently in the further text processing. Specifically, it allowed us to subject only the specific part of a court decision, in this case the court argumentation, to further the reference recognition and extraction. A textual segment recognised as the court argumentation is then processed further.

As the second step, parts recognised by the text segmentation model as a court argumentation was processed using the reference recognition model. After carefully studying the evaluation of the model's performance in \cite{harasta}, we have decided to use only part of the said model. Specifically, we have employed the recognition of the court identifiers, as we consider the rest of the smaller units introduced by Harašta and Šavelka of a lesser value for our task. Also, deploying only the recognition of the court identifiers allowed us to avoid the problematic parsing of smaller textual units into the references. The text spans recognised as identifiers of court decisions are then processed further.

At this point, it is necessary to evaluate the performance of the above mentioned part of the pipeline before proceeding further. The evaluation of the performance is summarised in Table \ref{tab:model}. It shows that organising the two models into the pipeline boosted the performance of the reference recognition model, leading to a higher F1 measure in the initial recognition of the text spans and their classification.

\begin{table}
 \caption{Model performance}
  \centering
  \begin{tabular}{ p{4 cm} | p{1 cm} | p{1 cm} | p{1 cm} | p{1 cm} | p{1 cm} | p{1 cm} }
    \multicolumn{1}{c|}{ } & \multicolumn{3}{|c|}{\textbf{Strict agreement}} & \multicolumn{3}{|c|}{\textbf{Overlap agreement}} \\
      \hline 
 \hline  & \textbf{P} & \textbf{R} & \textbf{F1} & \textbf{P} & \textbf{R} & \textbf{F1}  \\ \hline \hline 
    Reference recognition (court identifier) per \cite{harasta} & - & - & .652 & - & - & .709   \\ \hline 
    Text segmentation (argumentation detection) per \cite{harasta-segm}   & .885 & .950 & .915 & - & - & -   \\ \hline 
    Pipeline    & .732 & .716 & .724 & .846 & .786 & .815  \\ \hline 
    \bottomrule
  \end{tabular}
  \label{tab:model}
\end{table}

Further processing included:
\begin{enumerate}
    \item control and repair of incompletely identified court identifiers (manual);
    \item identification and sorting of identifiers as belonging to Supreme Court, Supreme Administrative Court or Constitutional Court (rule-based, manual);
    \item standardisation of different types of court identifiers (rule-based, manual);
        \item parsing of identifiers with court decisions available in CzCDC 1.0.
\end{enumerate}

\section{Results}
\label{sec:results}

Overall, through the process described in Section \ref{sec:methodology}, we have retrieved three datasets of extracted references - one dataset per each of the apex courts. These datasets consist of the individual pairs containing the identification of the decision from which the reference was retrieved, and the identification of the referred documents. As we only extracted references to other judicial decisions, we obtained 471,319 references from Supreme Court decisions, 167,237 references from Supreme Administrative Court decisions and 264,463 references from Constitutional Court Decisions. These are numbers of text spans identified as references prior the further processing described in Section \ref{sec:methodology}.

\begin{table}
 \caption{References sorted by categories, unlinked}
  \centering
  \begin{tabular}{lllll}
    \toprule
    \cmidrule(r){1-2}
    Court     & Supreme Court     & Supreme Adm. Court & Constitutional Court & Rest \\
    \midrule
    Supreme Court cites & 153 242  & 804 & 80 658  &112 287  \\
    Supreme Administrative Court cites   & 1 342 & 90 217 & 14 756 & 20 709  \\
    Constitutional Court cites    & 8 486   & 2 877   & 137 308 & 45 689  \\
    \bottomrule
  \end{tabular}
  \label{tab:stats-res}
\end{table}

These references include all identifiers extracted from the court decisions contained in the CzCDC 1.0. Therefore, this number includes all other court decisions, including lower courts, the Court of Justice of the European Union, the European Court of Human Rights, decisions of other public authorities etc. Therefore, it was necessary to classify these into references referring to decisions of the Supreme Court, Supreme Administrative Court, Constitutional Court and others. These groups then underwent a standardisation - or more precisely a resolution - of different court identifiers used by the Czech courts. Numbers of the references resulting from this step are shown in Table \ref{tab:stats-res}.

\begin{table}
 \caption{References linked with texts in CzCDC}
  \centering
  \begin{tabular}{llll}
    \toprule
    \cmidrule(r){1-2}
    Court     & Supreme Court     & Supreme Adm. Court &Constitutional Court \\
    \midrule
    Supreme Court cites & 140 355  & 658 & 76 003    \\
    Supreme Administrative Court cites   & 1 191 & 84 728 & 13 473      \\
    Constitutional Court cites    & 7 474  & 2 168   & 133 148 \\
    \bottomrule
  \end{tabular}
  \label{tab:stats-pars}
\end{table}

Following this step, we linked court identifiers with court decisions contained in the CzCDC 1.0. Given that, the CzCDC 1.0 does not contain all the decisions of the respective courts, we were not able to parse all the references. Numbers of the references resulting from this step are shown in Table \ref{tab:stats-pars}.

\section{Discussion}
\label{sec:discussion}
This paper introduced the first dataset of citation data of the three Czech apex courts. Understandably, there are some pitfalls and limitations to our approach.

As we admitted in the evaluation in Section \ref{pipeline}, the models we included in our NLP pipelines are far from perfect. Overall, we were able to achieve a reasonable recall and precision rate, which was further enhanced by several round of manual processing of the resulting data. However, it is safe to say that we did not manage to extract all the references. Similarly, because the CzCDC 1.0 dataset we used does not contain all the decisions of the respective courts, we were not able to parse all court identifiers to the documents these refer to. Therefore, the future work in this area may include further development of the resources we used. The CzCDC 1.0 would benefit from the inclusion of more documents of the Supreme Court, the reference recognition model would benefit from more refined training methods etc. 

That being said, the presented dataset is currently the only available resource of its kind focusing on the Czech court decisions that is freely available to research teams. This significantly reduces the costs necessary to conduct these types of studies involving network analysis, and the similar techniques requiring a large amount of citation data.

\section{Conclusion}
\label{sec:conclusion}
In this paper, we have described the process of the creation of the first dataset of citation data of the three Czech apex courts. The dataset is publicly available for download at \url{https://github.com/czech-case-law-relevance/czech-court-citations-dataset}.

\section*{Acknowledgment}
J.H., and T.N. gratefully acknowledge the support from the Czech Science Foundation under grant no. GA-17-20645S. T.N. also acknowledges the institutional support of the Masaryk University.
This paper was presented at CEILI Workshop on Legal Data Analysis held in conjunction with Jurix 2019 in Madrid, Spain.

\bibliographystyle{unsrt}  


\end{document}